\documentclass{WileyMSP-template}

\usepackage{amsmath} 
\usepackage{float}
\usepackage{caption}
\usepackage{ragged2e}
\usepackage{cite}

\begin{document}

\title{Diffusion Probabilistic Model Based Accurate and High-Degree-of-Freedom Metasurface Inverse Design}

\maketitle


\author{Zezhou Zhang}
\author{Chuanchuan Yang*}
\author{Yifeng Qin*}
\author{Hao Feng}
\author{Jiqiang Feng}
\author{Hongbin Li}

\dedication{ }

\setlength{\parindent}{1em}
\setlength{\parskip}{0.5em}
\captionsetup[figure]{name={Figure},labelfont=bf, labelsep=period}

\begin{affiliations}
Z. Zhang, H. Feng\\
Peking University Shenzhen Graduate School\\
Peking University, Shenzhen 518055, China\\
Email Address: zezhou.zhang@stu.pku.edu.cn

Z. Zhang, H. Feng, Prof. C. Yang, Prof. H. Li\\
The State Key Laboratory of Advanced Optical Communication Systems and Networks\\
School of Electronics, Peking University, Beijing 100871, China\\
Email Address: yangchuanchuan@pku.edu.cn

Z. Zhang, Prof. C. Yang, Dr. Y. Qin, H. Feng, Prof. J. Feng, Prof. H. Li \\
Peng Cheng Laboratory, Shenzhen 518055, China\\
Email Address: qinyf@pcl.ac.cn

Prof. J. Feng\\
College of Mathematics and Statistics, Shenzhen University, Shenzhen 518060, China

\end{affiliations}

\keywords{deep learning, metasurfaces, inverse design, diffusion probabilistic model}

\justifying

\begin{abstract}
Conventional meta-atom designs rely heavily on researchers’ prior knowledge and trial-and-error searches using full-wave simulations, resulting in time-consuming and inefficient processes. Inverse design methods based on optimization algorithms, such as evolutionary algorithms, and topological optimizations, have been introduced to design metamaterials. However, none of these algorithms are general enough to fulfill multi-objective tasks. Recently, deep learning methods represented by Generative Adversarial Networks (GANs) have been applied to inverse design of metamaterials, which can directly generate high-degree-of-freedom meta-atoms based on S-parameter requirements. However, the adversarial training process of GANs makes the network unstable and results in high modeling costs. This paper proposes a novel metamaterial inverse design method based on the diffusion probability theory. By learning the Markov process that transforms the original structure into a Gaussian distribution, the proposed method can gradually remove the noise starting from the Gaussian distribution and generate new high-degree-of-freedom meta-atoms that meet S-parameter conditions, which avoids the model instability introduced by the adversarial training process of GANs and ensures more accurate and high-quality generation results. Experiments have proven that our method is superior to representative methods of GANs in terms of model convergence speed, generation accuracy, and quality.
\end{abstract}



\section{Introduction}
The prevalent metamaterial/functional-metasurface design involves trial-and-error iterative process guided by designers' intuition, that designers select unit cell structures, perform parameter sweeps, and create arrays to achieve specific functions.\textsuperscript{\cite{ref1,ref2,ref3,ref4}} However, this method is inefficient and struggles with wideband multi-frequency targets due to increased nonlinearity.\textsuperscript{\cite{ref5}} Recently, optimization-based inverse design methods, such as adjoint-based topological optimization,\textsuperscript{\cite{ref6}} genetic algorithm,\textsuperscript{\cite{ref7}} and ant-colony optimization,\textsuperscript{\cite{ref8}} have emerged, but they cannot deal with multi-objective tasks and have high computational costs due to individual optimization for different targeted unit cells.\par

Compared to optimization-based inverse design methods, deep learning can effectively explore the global design space,\textsuperscript{\cite{ref9}} enabling fast generation of meta-atoms’ structures meeting S-parameter requirements that can be reused after a single training. Early research employed a tandem approach to connect forward and inverse networks to solve one-to-many problems, using fixed structure parameters in one-dimensional vectors for inverse design.\textsuperscript{\cite{ref10,tandem1,tandem2,tandem3,tandem4}} As performance requirements grew, researchers explored free-form structures with higher degrees of freedom, and introduced Generative Adversarial Network(GAN)\textsuperscript{\cite{ref11}} and Variational Autoencoder(VAE)\textsuperscript{\cite{ref12}} as generation methods. \par

VAEs require additional optimization algorithms for extensive latent space searches {\cite{VAE1,VAE2,VAE3}}, increasing design complexity and producing blurry, low-quality results due to information loss during compression and reconstruction.\textsuperscript{\cite{ref5,ref13}} GANs, through adversarial training between a generator and a discriminator,\textsuperscript{\cite{ref14}} enable the direct generation of high-quality structures. Conditional GANs guide the network by providing conditional inputs, and have been applied to various scenarios like metasurfaces,\textsuperscript{\cite{ref15, so2019designing, wang2020gan, yeung2021global}} 1D meta-gratings,\textsuperscript{\cite{ref9}} transmission surfaces,\textsuperscript{\cite{ref16}} and single-frequency XY-polarization multiplexing surfaces.\textsuperscript{\cite{ref17}}  \par

However, GANs have inherent convergence difficulties during training, impacting their ability to generate accurate structures. Despite attempts to alleviate instability through methods like Deep Convolutional GANs (DCGANs),\textsuperscript{\cite{ref18}} Wasserstein GANs (WGANs),\textsuperscript{\cite{ref19}} and WGANs with Gradient Penalty (WGAN-GP),\textsuperscript{\cite{ref20}} the adversarial process's inherent instability remains unresolved. Model stability and accuracy still rely on the careful model structure design and hyperparameter selection during training.\textsuperscript{\cite{ref21}} When input and output dimensions change, numerous experiments are needed to adjust the model structure, increasing network design costs and limiting GANs' applicability in various scenarios.\textsuperscript{\cite{ref22}}\par

Recently, diffusion probability theory\textsuperscript{\cite{ref23}} has demonstrated superior performance over GANs in terms of accuracy, diversity, and precision for generated samples in image generation tasks.\textsuperscript{\cite{ref22}} Consequently, it has been successfully applied to text-to-image generation, including OpenAI's DALLE,\textsuperscript{\cite{ref24}} Google's IMAGEN,\textsuperscript{\cite{ref25}} and image super-resolution.\textsuperscript{\cite{ref26}} However, the application of diffusion probability theory as a novel generation method for on-demand direct inverse generation of metasurfaces remains unexplored.\par

\begin{figure}[t]
  \begin{center}
  \includegraphics[width=0.9\linewidth]{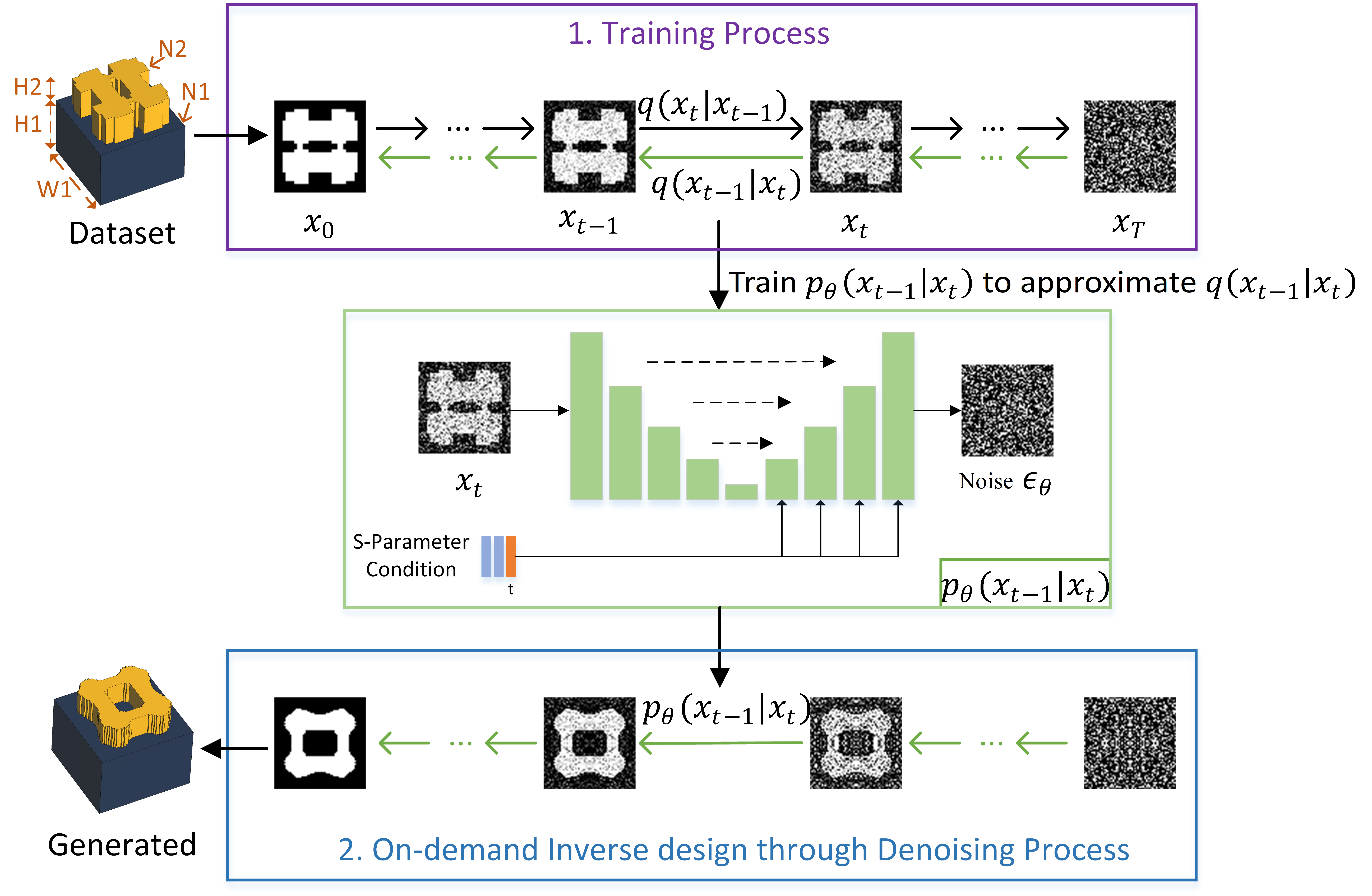}
  \end{center}
  \caption{Overview of MetaDiffusion. (a)Purple Panel: Schematic diagram of the training process. From left to right, the process of gradually adding noise to the original meta-atom is shown (black arrow), while from right to left, the process of denoising by gradually removing noise from samples on a Gaussian distribution to regenerate meta-atoms is shown (green arrow). (b)Green Panel: A neural network is used to approximate the posterior probability $q(x_{t-1}  |  x_t)$ required in the denoising process. By inputting the noisy meta-atom $x_t$, the noise $\epsilon_{\theta}$  to be removed is predicted. The S-parameter, extra parameters, and time step $t$ are used as conditions and are integrated into the neural network after feature extraction to control conditional denoising. (c)Blue Panel: On-demand Inverse design Process. The well-trained Neural networks is used in denoising process to generate new meta-atoms that meet the requirements, by starting with random noise and performing directional denoising according to the S-parameter conditions. }
  \label{fig:metadiff_schem}
\end{figure}

In this paper, we propose a novel metasurface inverse design method based on diffusion probabilistic model, called MetaDiffusion, capable of directly generating free-form meta-atom structures with high degrees of freedom according to broadband amplitude and phase requirements. Our method, illustrated in \textbf{Figure \ref{fig:metadiff_schem}}, initiates a forward Markov process by gradually adding noise to the meta-atom $x_0$ until it becomes Gaussian noise $x_T$ in $q(x_t   |  x_{t-1})$. Then, a neural network $p_\theta (x_{t-1}  |  x_t)$ approximates the denoising process $q(x_{t-1}  |  x_t)$, iteratively denoising the sampled $x_T$ in the Gaussian distribution to generate a new structure $x_0^{'}$ meeting the S-parameter requirements. This eliminates the unstable adversarial process employed by GANs. Consequently, MetaDiffusion bypasses the need of fine-tuning stable networks and hyperparameters for different data, and can be easily applied to various metasurface design tasks. We also fuse the S-parameters and extra parameters (meta-atom size $W1$, thickness $H2$, material refractive index $N2$) as a condition in the decoder of our model and utilize a classifier-free condition control strategy,\textsuperscript{\cite{ref27}} enabling accurate direct generation of wideband S-parameter targets.\par

 To the best of our knowledge, we are the first to apply the diffusion probabilistic model to inverse design of meta-atoms, overcoming the instability of GANs' adversarial process. Our method, MetaDiffusion, incorporates S-parameter conditions, and utilizes classifier-free condition control strategies in the neural network to achieve precise generation according to S-parameter requirements. Importantly, through rigorous tests, we demonstrate that our proposed method converges faster, exhibits higher conditional accuracy, and is more stable compared to GAN-based methods such as SLMGAN\textsuperscript{\cite{ref28}}  and WGAN-GP. Specifically, for a wideband S-parameter target with 52 sampling points between 30-60THz, our method shows a $43\%$ and $48\%$ improvement in average mean absolute error of the S-parameter compared to SLMGAN and WGAN-GP, respectively. Furthermore, the mean absolute errors of top $95\%$ samples are below 0.071, which is $38\%$ better than SLMGAN and $49\%$ better than WGAN-GP.\par

\newpage
\section{The Proposed MetaDiffusion Model}
In order to train the neural network to gradually denoise Gaussian noise and generate a new meta-atom according to the S-parameter requirements, noise is first added to the meta-atoms in the training set through a forward Markov process. Starting from the original meta-atom $x_0$, random noise $\epsilon$ is added at each time step $t\in{1,2,\dots,T}$  according to the variance scheduler $\beta_1,\dots,\beta_T$:
\begin{align}
    q\left(x_{t}  |  x_{t-1}\right) &:= \mathcal{N}\left(x_{t} ; \sqrt{1-\beta_{t}} x_{t-1}, \beta_{t} \mathbf{I}\right), \label{eq1}\\
    q\left(x_{1: T}  |  x_{0}\right) &:= \prod_{t=1}^{T} q\left(x_{t}  |  x_{t-1}\right), \label{eq2}
\end{align}
where the variance $\beta_t \in (0, 0.02)$ increases linearly from the first timestep $\beta_1 = 1e-4$ to the last timestep $\beta_T=0.02$, indicating that more noise is gradually added to the data so that the image eventually approaches a Gaussian distribution. By using reparameterization trick and defining $\alpha_t=1-\beta_t$, ${\bar{\alpha}}_t=\prod_{s=1}^{t}\alpha_s$, the noisy meta-atom matrix $x_t$ at any time step $t$ can be directly expressed as a closed-form formula based on $x_0$ (detailed formula derivation see section 1.1 of the Supporting Information):
\begin{align}
    x_{t}=\sqrt{\bar{\alpha}_{t}} x_{0}+\epsilon \sqrt{1-\bar{\alpha}_{t}}, \epsilon \sim \mathcal{N}(0,1). \label{eq3}
\end{align}

To gradually remove noise from the samples $x_T$ in the Gaussian distribution and obtain $x_0$ from the original meta-atom distribution, the posterior probability $q(x_{t-1} |  x_t)$ need to be known, which is intractable. Therefore, we use a learnable function $p_\theta (x_{t-1} |  x_t)$ to approximate $q(x_{t-1} |  x_t)$ instead. The denoising process from $x_T$ to $x_0$ is defined as $p_\theta(x_{0:T})$:
\begin{align}
p_{\theta}\left(x_{0: T}\right) &=p\left(x_{T}\right) \prod_{t=1}^{T} p_{\theta}\left(x_{t-1}  |  x_{t}\right), \\
p_{\theta}\left(x_{t-1}  |  x_{t}\right)&=\mathcal{N}\left(x_{t-1} ; \ \mu_{\theta}\left(x_{t}, t\right), {{\Sigma }_{\theta }}({{x}_{t}},t)\right).
\end{align}

During the training process, we encourage $p_\theta (x_{t-1}  |  x_t)$ to approximate $q(x_{t-1}  |  x_t)$ by optimizing the variational lower bound on negative log likelihood, where variational lower bound $L$ can be simplified as (detailed formula derivation see section 1.2 of the Supporting Information):
\begin{align}
E\left[-\log p_{\theta}\left(x_{0}\right)\right] \leq L &:=E_{q}\left[-\log \frac{p_{\theta}\left(x_{0: T}\right)}{q\left(x_{1: T}  |  x_{0}\right)}\right], \label{eq6}\\
& \propto \sum_{t>1} D_{\mathrm{KL}}\left(q\left(x_{t-1}  |  x_{t}, x_{0}\right) \|   p_{\theta}\left(x_{t-1}  |  x_{t}\right)\right). \label{eq7}
\end{align}

Minimizing the KL divergence can be intuitively understood as using the neural network $\epsilon_\theta (x_t,t)$ to approximate the estimation of the noise $\epsilon_t$ added at each time step in the forward process (detailed formula derivation see section 1.3 of the Supporting Information),
\begin{align}
     \mathop{\arg \min}\limits_\theta L(\theta) \propto \mathop{\arg \min}\limits_\theta E\left[\left\|\epsilon_{t}-\epsilon_{\theta}\left(x_{t}, t\right)\right\|^{2}\right], \label{eq8}
\end{align}
where $\theta$ represents the learnable parameters of a neural network, and $x_t$ can be obtained from Equation \ref{eq3}.

For the purpose of incorporating the S-parameter and extra parameters (meta-atom size $W1$, thickness $H2$, material refractive index $N2$) to control the generation process of the model and ensure that the generated structure with specific material, thickness, and size, meets the requirements of the S-parameter condition, the conditions are fused with time information $t$ and used as part of the neural network input during training process, represented as $\epsilon_\theta (x_t,t,c)$. In addition, in order to further improve the accuracy of the model, we adopt a classifier-free guide strategy, which trained both a conditional denoising diffusion model and an unconditional denoising diffusion model on the neural network simultaneously as an implicit guide. Specifically, during training, $10\%$ of data samples are randomly selected, and their corresponding conditions are masked with a specific label 0. The complete training process can be found in Algorithm S1 in the Supporting Information.

After the neural network is trained, during the on-demand generation stage of the meta-atom, we mix the conditional estimate $\epsilon_\theta (x_t,t,c)$ and unconditional estimate $\epsilon_\theta (x_t,t,0)$ as the network’s noise prediction result, where $w$ is a hyperparameter that controls the trade-off between accuracy and diversity.
\begin{align}
\tilde{\epsilon}_{\theta}\left(x_{t}, t, c\right)=(1+w) \epsilon_{\theta}\left(x_{t}, t, c\right)-w \epsilon_{\theta}\left(x_{t}, t, 0\right).
\end{align}

Then, we use the reparameterization trick on the posterior $q(x_{t-1} |  x_t,x_0 )$ and substitute the noise $\Tilde{\epsilon}_\theta (x_t,t,c)$ approximated by the neural network to obtain an image from the former time step (detailed formula derivation see section 1.4 of the Supporting Information):
\begin{align}
    x_{t-1}=\frac{1}{\sqrt{\alpha_{t}}}\left(x_{t}-\frac{1-\alpha_{t}}{\sqrt{1-\bar{\alpha}_{t}}} \tilde{\epsilon}_{\theta}\left(x_{t}, t, c\right)\right)+\frac{1-{{{\bar{\alpha }}}_{t-1}}}{1-{{{\bar{\alpha }}}_{t}}}\cdot {{\beta }_{t}}\cdot z , z \sim \mathcal{N}(0,1).
\end{align}

 Through the iterative denoising process, new meta-atoms that meet the requirements of the S-parameter can be generated. The complete process of the on-demand generation stage can be found in Algorithm S2 in the Supporting Information.

\begin{figure}[H]
  \includegraphics[width=\linewidth]{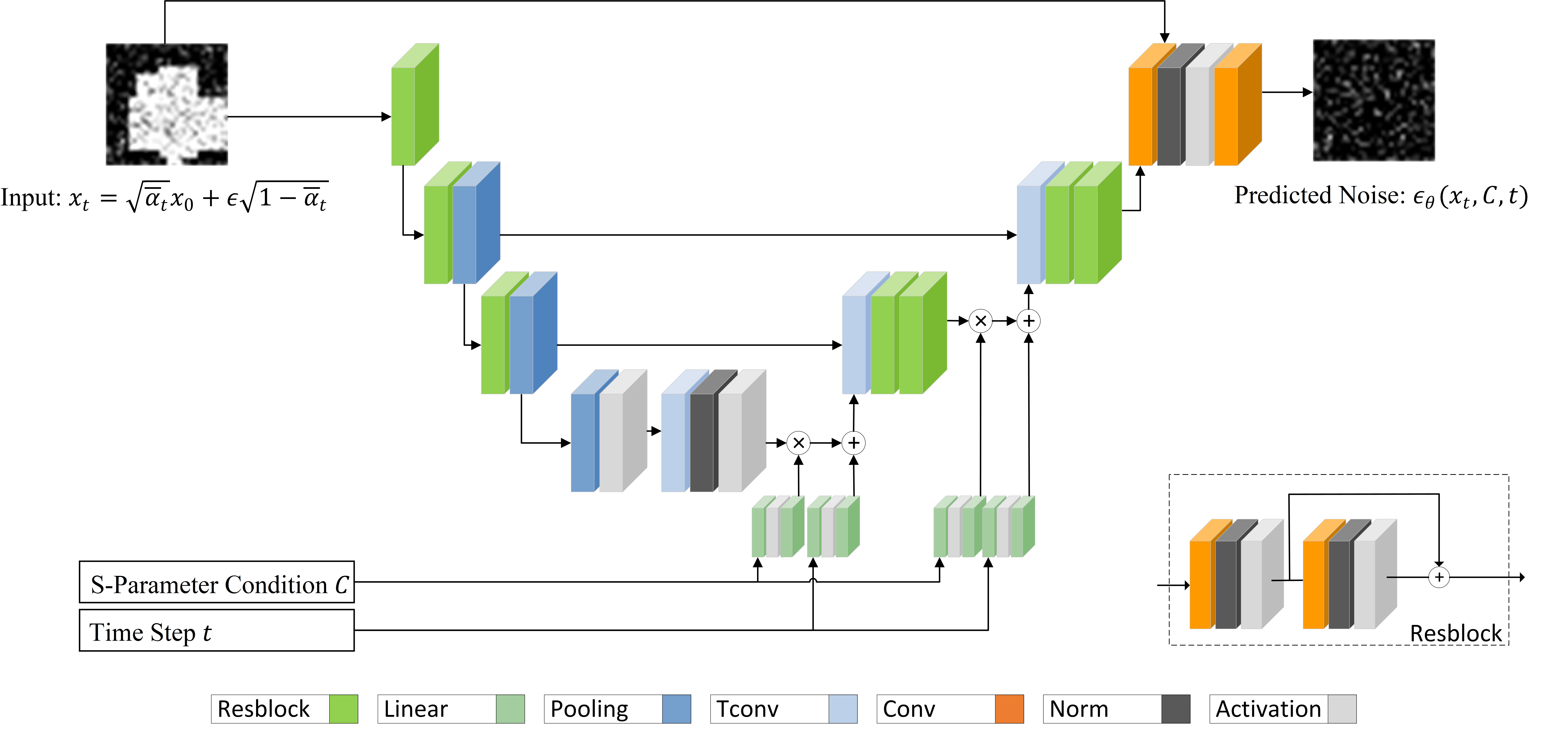}
  \caption{MetaDiffusion network architecture. The input $x_t$ is first downsampled by an encoder composed of residual blocks  (Resblock) and pooling layers, reducing the matrix size while increasing the depth of the feature channels. At the bottleneck layer, it becomes a 512-length vector. It then passes through a decoder composed of multiple transposed convolutional layers  (Tconv) and residual blocks, gradually reducing the number of channels and restoring the size. There are skip connections between the encoder and decoder to retain information. The S-parameters and extra parameters are concatenated to a condition vector. The condition vector and time step information are separately passed through feature extractors composed of fully connected layers (Linear) before being added to the decoder. Finally, the network predict noise $\epsilon_\theta$ of the same size as the input.}
  \label{fig:network_achitecture}
\end{figure}

As shown in \textbf{Figure \ref{fig:network_achitecture}}, we employ a generic encoder-decoder U-Net architecture for predicting noise $\epsilon$ for a given noisy image $x_t$. This enable us to approximate the posterior probability $q(x_{t-1}|x_t)$ and subsequently obtain $x_{t-1}$ using Equation 10. The network consists of a set of encoders composed of residual blocks and pooling layers for down sampling compression and then uses transposed convolutions and residual blocks for up sampling recovery. Key features are extracted during the compression and recovery process. The encoder and decoder are connected by skip connections to alleviate information loss during the down sampling process of the encoder. In addition, after extracting features from the S-parameter, extra parameters and time step $t$, we fused the embedded features into the up-sampling process of the decoder so that the S-parameter, extra parameters and time step information could guide the denoising process of the network as conditions. (For specific network structure and training hyperparameters, see the Supporting Information)

\section{Results and Discussion}
To verify and compare the performance of the model, we use the all-dielectric freeform unit cell dataset.\textsuperscript{\cite{ref30}} As shown in Fig.1, each unit cell consists of two layers of dielectric with different refractive indices, where the upper layer is a high-refractive-index free-form structure symmetrical about the XY axis and the lower layer is a fixed refractive index of $N1=1.4$ and a thickness of $H1=2 \ \mu m$. The entire unit cell structure can be characterized by a $64\times 64$ binary matrix (0 represents air, 1 represents dielectric) to represent the shape of the free-form structure and a $1\times3$ parameter matrix to represent the substrate size $W1\in[2.5\ \mu m,3\ \mu m]$, thickness $H2\in[0.5\ \mu m,1\ \mu m]$, and refractive index $N2\in[3.5,5]$. The dataset assumes that all meta-atoms are polarization-independent and reciprocal, with symmetric configurations. This allow us to simplify the structure representation to a $32\times 32$ two-dimensional matrix in the upper left corner, thus introducing physical symmetry as prior knowledge in advance, which is conducive to network training. However, in the future, we plan to use our MetaDiffusion model to generate meta-atoms with more sophisticated properties, such as chiral metamaterials or non-Hermitian metamaterials. The transmission responses from $30$ to $60$ THz obtained by simulating the structure using CST Microwave Studio are used as the label for the structure data. The real and imaginary parts of the transmission response are sampled at $26$ points respectively and combined into a $1\times52$ one-dimensional vector to represent the transmission response. In addition, we incorporate $1\times3$ structural parameters$[W1,H2,N2]$ as extra conditions, thus adding a constraint that guides the model to generate meta-atoms in the desired material and structural size and thickness. As a result, we concatenate the $1\times3$ parameter vector with the $1\times52$ transmission response to form a $1\times55$ vector as the target condition for generating network input. The dataset contains a total of 174883 data points and is divided into training, validation, and test sets at a ratio of $8:1:1$.

During the process of training and comparing generative neural networks, the model is instructed to generate structures for the S-parameter targets on the validation and test sets. And then, the generated results need to be forward simulated to obtain the corresponding frequency response, thereby quantitatively evaluating the accuracy performance of the network. However, using numerical simulation methods such as Finite-Difference Time-Domain(FDTD) is very time-consuming and much slower than the generation speed of the generative network. It is difficult to handle the evaluation of large amounts of data during the training and testing process of neural networks. Therefore, we choose to use a Predicting Neural Network (PNN)\textsuperscript{\cite{ref30}} as a fast solver instead to accurately predict the spectral response of unit cells within milliseconds. We have verified through qualitative and quantitative experiments that PNN can serve as a reliable surrogate solver. (See the Supporting Information for experimental results)

To fully demonstrate the accuracy and stability of the on-demand direct generation capabilities of the method proposed in this paper, we compare it with the current state-of-the-art methods for on-demand direct generation of meta-atoms, including WGAN-GP and SLMGAN with a simulator. (1) WGAN-GP enhances the stability of WGAN by employing a gradient penalty method to restrict the parameters of the network that approximates the Wasserstein distance. (2) The recently proposed SLMGAN with a simulator (hereinafter referred to as SLMGAN) further enhances training stability and conditional accuracy by employing Sinkorn iteration to explicitly calculate optimal transport and incorporating a pre-trained forward prediction network as one of the training loss items during training. (See the Supporting Information for implementation details of SLMGAN with Simulator and WGAN-GP)

After binarizing the generated results of the three models (using 0.5 as the threshold, setting values less than 0.5 to 0 and values greater than or equal to 0.5 to 1), we use the surrogate solver for forward calculation on them to obtain the corresponding S-parameters. Mean Absolute Error (MAE) is used as the error indicator for the accuracy of each sample, where MAE of each sample is expressed as:
\begin{align*}
    L=\dfrac{1}{n}\sum_{i=1}^n|s_{target}(i)-s_{gen}(i)|,
\end{align*}
where $s_{target}(i)$ is the target S-parameter at the $i$-th frequency point and $s_{gen}(i)$ is the S-parameter corresponding to the generated result at the $i$-th frequency point. The above calculation includes 26 points for both the real part and the imaginary part.

\begin{figure}[H]
\begin{center}
      \includegraphics[width=0.9\linewidth]{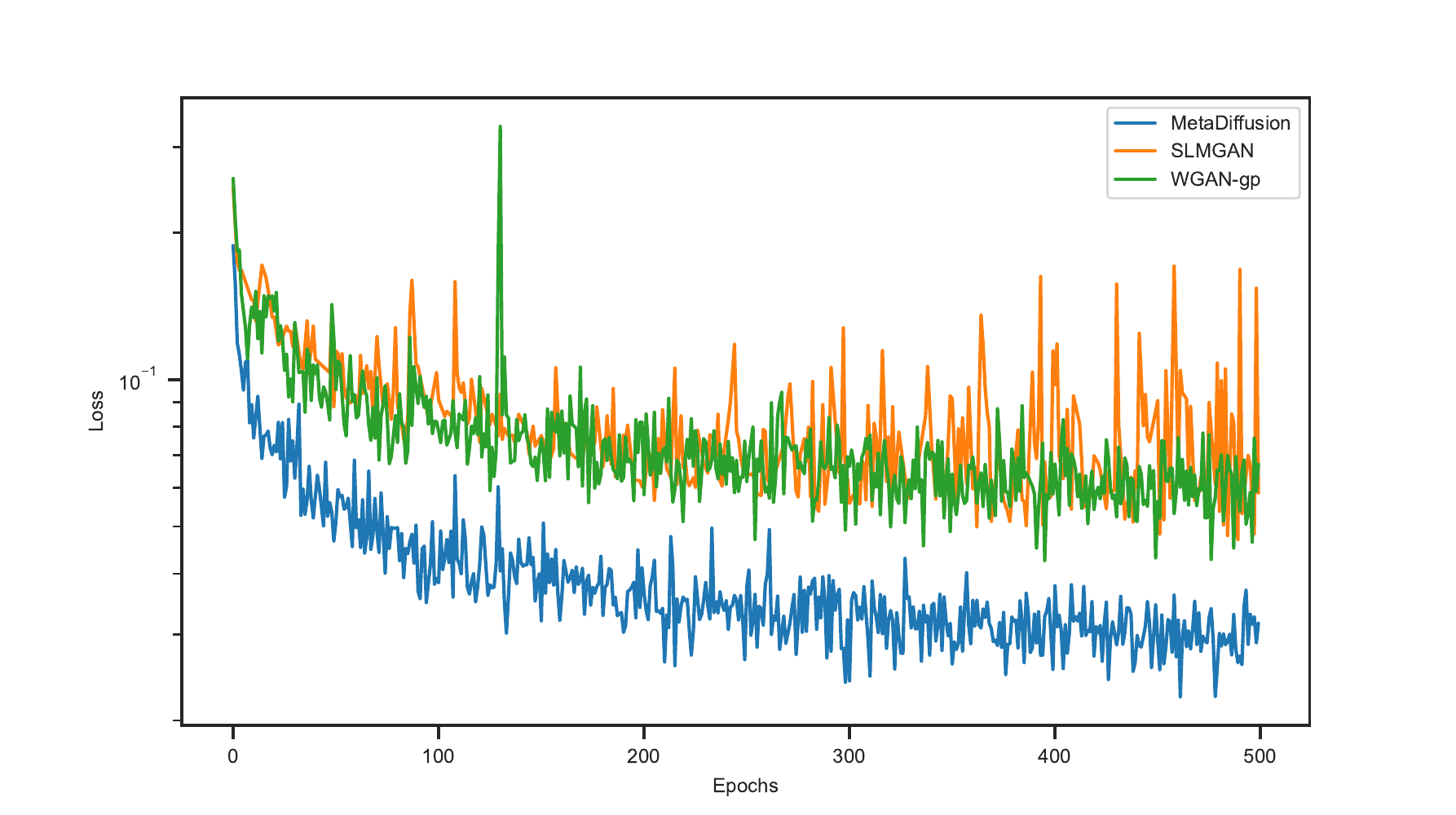}
\end{center}
  \caption{Validation loss of three models during the training process.}
  \label{fig:fig3}
\end{figure}

During the training process, at the end of each epoch, we use the model parameters of the current epoch to generate samples on the validation set according to the S-parameter conditions and evaluate the MAE of each generated sample relative to the input conditions. The mean of the MAE loss for all samples are calculated to obtain the accuracy of the generated results on the entire validation set under the current epoch. The MAE results of the three methods on the validation set during the entire training process are shown in \textbf{Figure \ref{fig:fig3}}. All three methods gradually converge within 500 epochs, with MetaDiffusion converging to around 0.03 and SLMGAN and WGAN-GP methods converging to around 0.05 and 0.06 respectively. More importantly, from the figure we can also see that our method converges significantly faster than SLMGAN and WGAN-GP. Specifically, MetaDiffusion converged to 0.068 in the first 14 epochs and to 0.052 in the first 30 epochs, indicating that our method is more stable and easier to train than GAN-based methods, and requires less training time to achieve the same effect.

Then, the three trained models is used to perform qualitative and quantitative comparisons of model performance on the test set. In \textbf{Figure 4-5}, the generation results of the three models are randomly selected and qualitatively presented. Each row represents one method, from top to bottom: MetaDiffusion, SLMGAN, and WGAN-GP. For each method, the figure shows the original structure in the dataset, the direct output of the model, and the structure after binarization. Finally, the binarized structure is forward-simulated to compare the S-parameters of the generated structure with the target S-parameters. From the figure, we can see that all three methods can generate structures that roughly conform to the target S-parameter trend according to the required S-parameter conditions. However, our proposed MetaDiffusion method has the highest accuracy among the three methods, while SLMGAN and WGAN-GP show inaccurate phenomena in some frequency ranges. Although we use the results after binarization as the outputs of the models, directly observing the outputs of the models can reflect the generation ability and stability of the model to a certain extent. From the figure, it can be seen that MetaDiffusion can directly output clear structural images, while WGAN-GP has some blurring phenomena in its generated results due to training instability caused by adversarial processes. SLMGAN further affects the clarity of its generated results due to its use of special interpolation operations as a way to introduce symmetry information. 

\begin{figure}[H]
  \begin{center}
  \includegraphics[width=\linewidth]{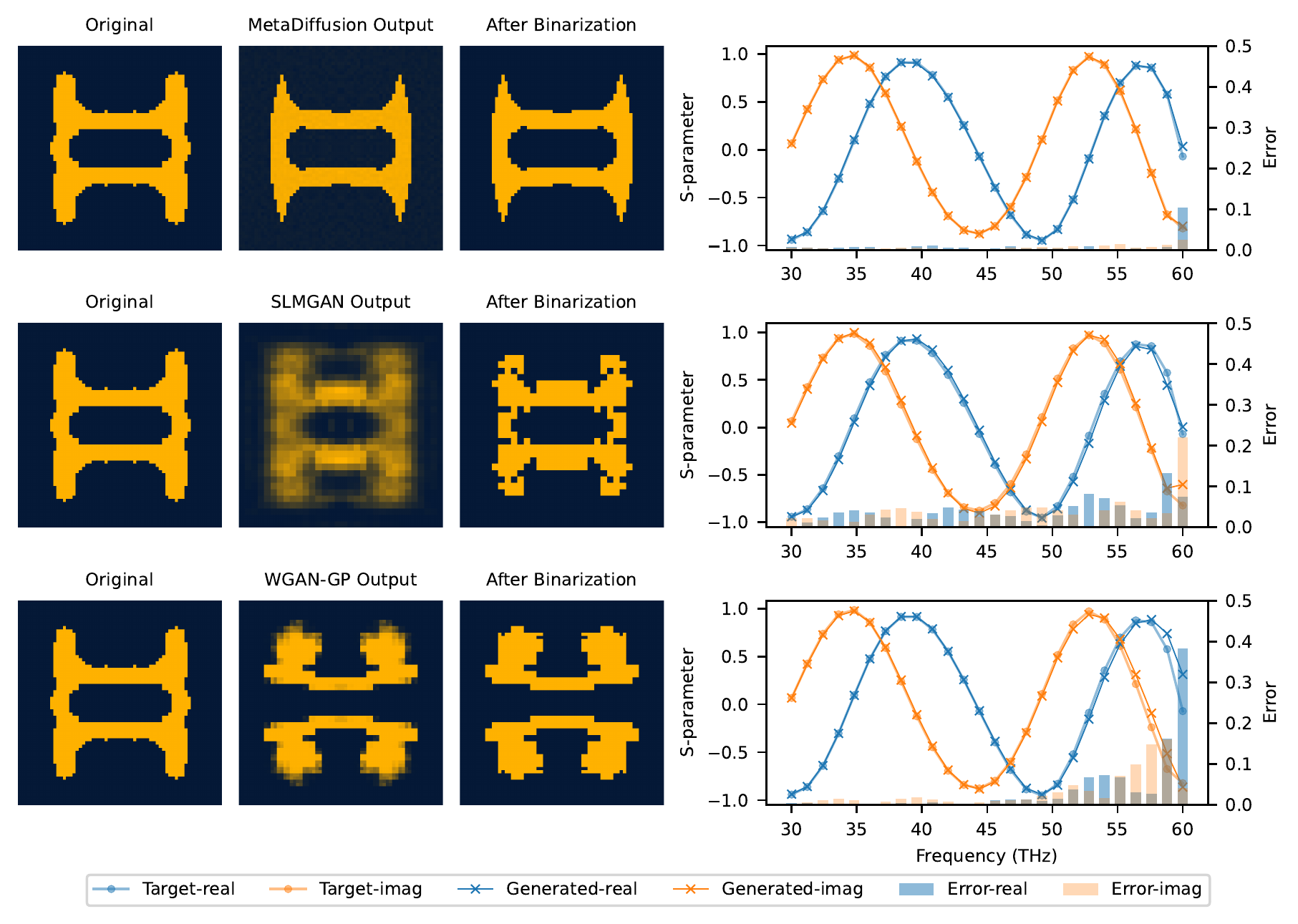}
  \end{center}
  \caption{Qualitative comparison example 1 of on-demand inverse design. From top to bottom, the three rows show the generated results of MetaDiffusion, SLMGAN with simulation, and WGAN-GP. For each method, we display the structure in the original dataset (Original), the direct output of the model (Model Output), and the output after binarization (After Binarization). We perform forward simulation on the binarized structure to obtain the real part (Generated-real) and imaginary part (Generated-imag) of the corresponding S-parameter and compare it with the real part (Target-real) and imaginary part (Target-imag) of the target S-parameter. The error in the real part (Error-real) and imaginary part (Error-imag) at each frequency point is shown as a bar graph. Note that the gray portion of the figure represents the overlap of blue and orange.}
  \label{fig:fig4}
\end{figure}

\begin{figure}[H]
  \begin{center}
  \includegraphics[width=\linewidth]{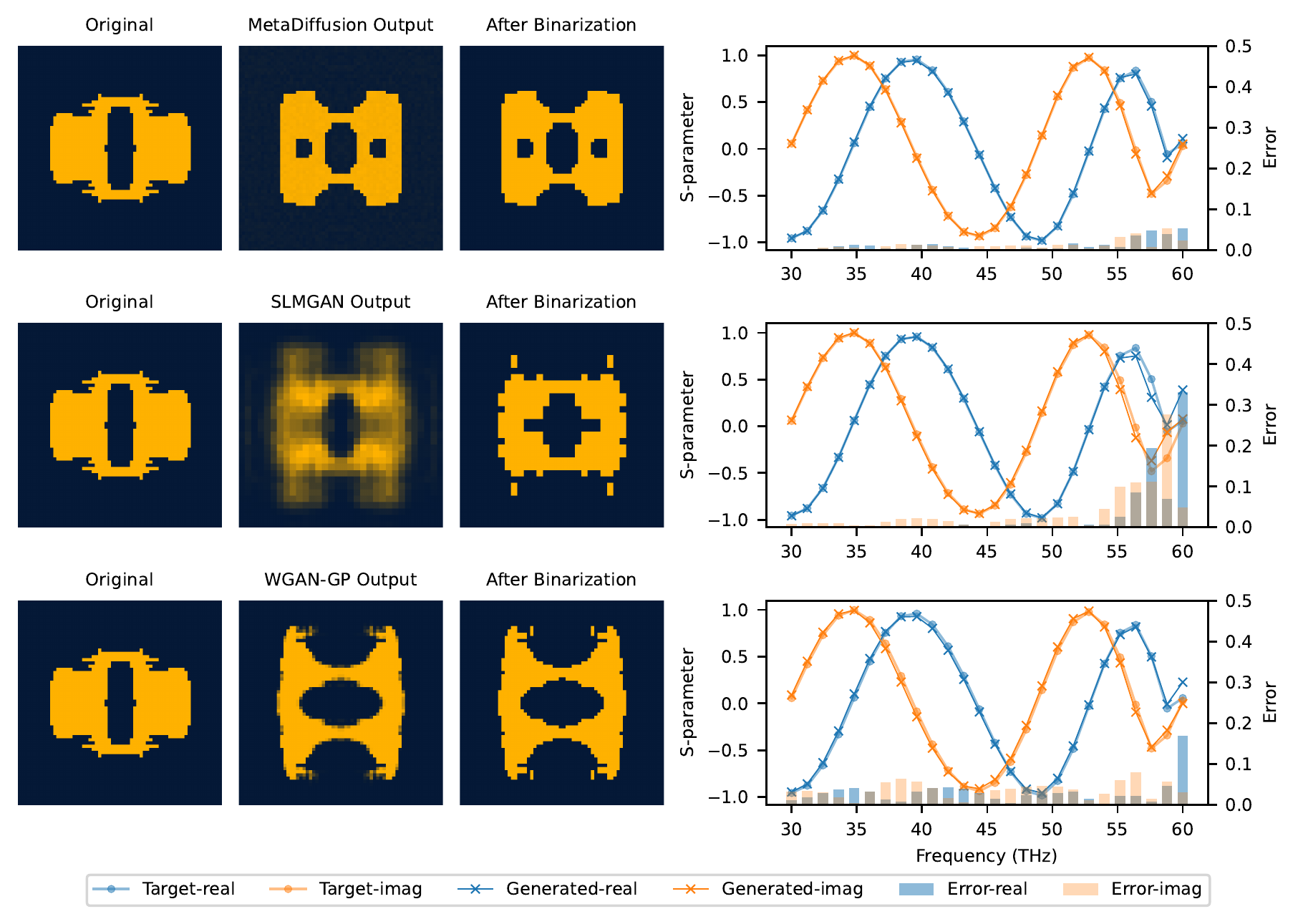}
  \end{center}
  \caption{Qualitative comparison example 2 of on-demand inverse design.}
  \label{fig:fig5}
\end{figure}

Additionally, as depicted in Figures 4-5, our proposed method tends to produce more uniform structures, whereas the structures generated by the other two methods occasionally exhibit scattered pixels, complicating the fabrication process. To further demonstrate the advantage of our method in generating stable structures with respect to ease of fabrication, we employ our approach to perform one-to-many generations for specific targets, as illustrated in \textbf{Figure 6}.

\begin{figure}[H]
  \includegraphics[width=\linewidth]{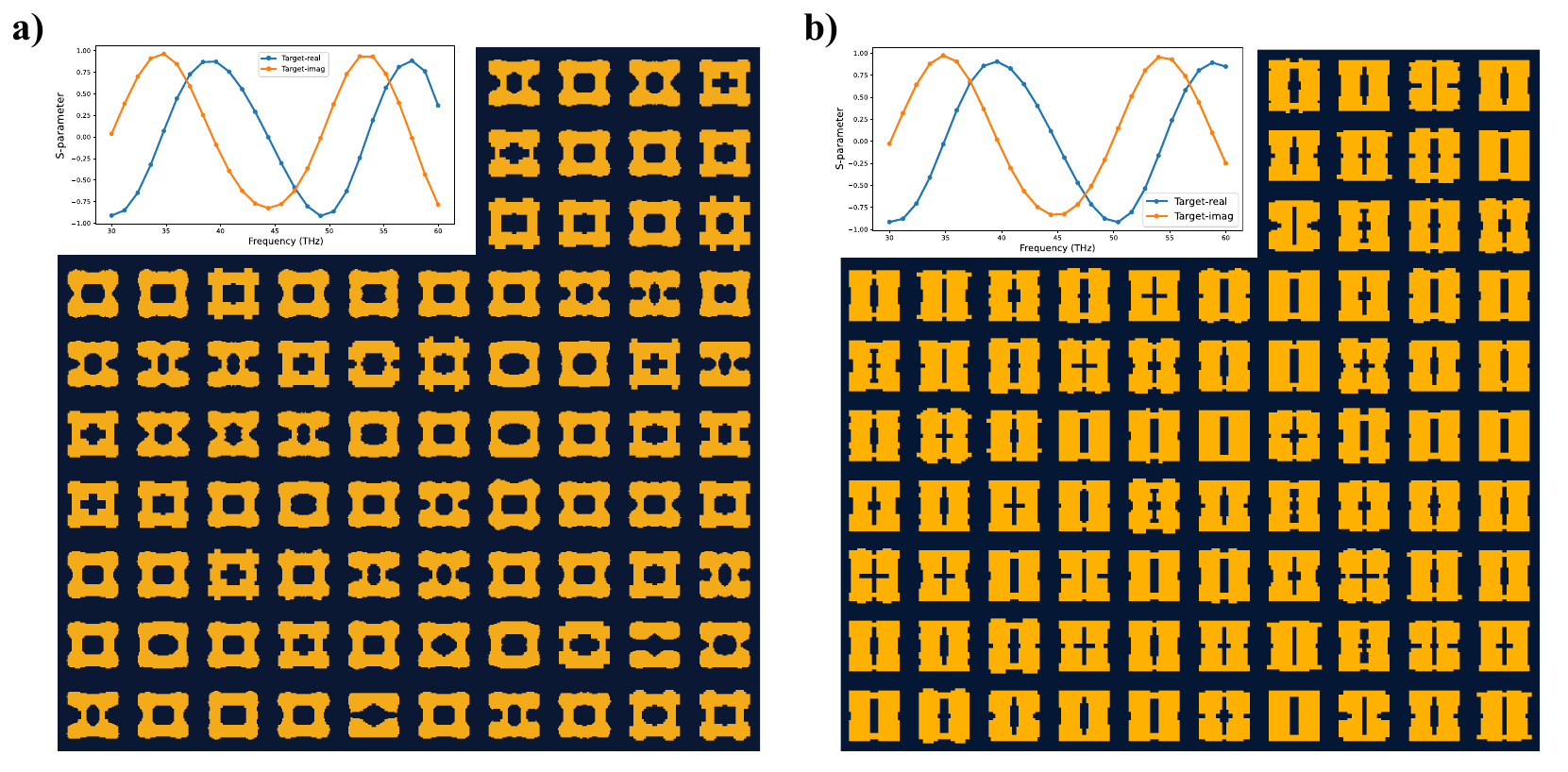}
  \caption{One to many Generations. Panels a) and b) show two sets of examples where multiple unit cell structures that meet the given S-parameter target (top left corner of each panel) are generated. It can be observed that MetaDiffusion tends to generate regular structures that are easy to fabrication.}
  \label{fig:fig6}
\end{figure}

\begin{figure}[H]
  \includegraphics[width=\linewidth]{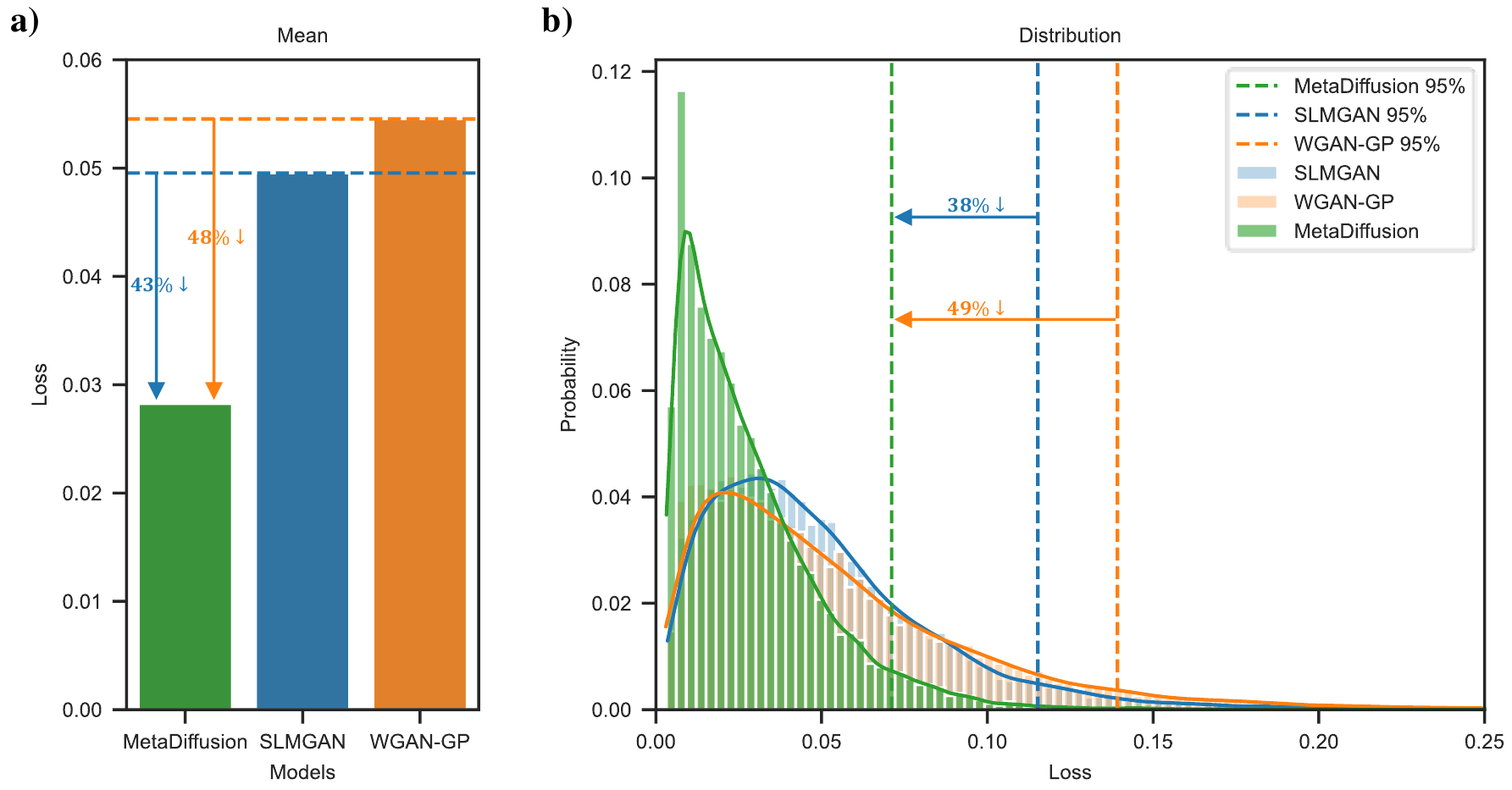}
  \caption{Quantitative comparison of the three methods. (a) Mean MAE error on the test set. The improvement of MetaDiffusion relative to SLMGAN with a simulator and WGAN-GP is indicated by arrows in the figure. (b) Error distribution on the test set. The vertical dashed line indicates the position corresponding to $95\%$ error.}
  \label{fig:fig7}
\end{figure}

\textbf{Figure 7} quantitatively compares the accuracy of the three models in generating unit cells according to S-parameter conditions on the entire test set. Figure 7-a shows the mean MAE on the entire test set. The average error of MetaDiffusion on the entire test set is 0.02824, while that of SLMGAN is 0.04955 and that of WGAN-GP is 0.05454. Therefore, our method has a $43\%$ and $48\%$ accuracy improvement compared to SLMGAN and WGAN-GP respectively.

In addition to focusing on the average error situation, we also pay attention to the distribution of errors in the samples to prevent situations where some samples performed extremely well and masked the poor performance of most samples in the mean indicator. We calculate the error of each individual sample for the three models on the entire test set. As shown in the error distribution graph in Figure 7-b, the green, blue, and orange bars represent the proportions of MetaDiffusion, SLMGAN, and WGAN-GP in the corresponding error range; the solid lines represent the use of kernel density estimate (KDE) data smoothing method on the corresponding model distribution to visualize the overall trend of the distribution. From Figure 7-b, it can be seen that the error distribution of our proposed method is more concentrated towards 0, while SLMGAN and WGAN-GP have a more even overall distribution. We mark with a dashed line in the figure the position of $95\%$ sample error for each model. In our proposed MetaDiffusion model, $95\%$ of samples are better than 0.071, while SLMGAN and WGAN-GP have $95\%$ errors of 0.115 and 0.139 respectively. Therefore, our method improves by $38\%$ and $49\%$ respectively compared to SLMGAN and WGAN-GP in terms of error for the top $95\%$ samples.

\section{Conclusion}

In conclusion, we present a novel inverse design method for metasurfaces, MetaDiffusion, based on diffusion probability theory. This method facilitates the direct generation of high-quality, diverse, and accurate freeform meta-atoms, conforming to broadband amplitude and phase requirements. By employing a neural network to learn the noise diffusion process, our approach generates new meta-atoms that meet S-parameter conditions via denoising, thereby avoiding the labor-intensive traditional iterative search process.

Compared to widely-used GAN series methods in recent AI applications for on-demand direct meta-atoms generation, our method circumvents the model instability introduced by the adversarial training process in GAN theory. Consequently, our approach obviates the need for multiple experiments to select stable network structures for different scenarios, allowing for easy extension to various situations while ensuring accurate and high-quality meta-atoms generation results. Experimental validation demonstrates that MetaDiffusion outperforms representative GAN methods in terms of model convergence speed, quality, and accuracy of generated structures.

The methodology, MetaDiffusion, establishes a novel direction in inverse design and fosters further inquiry into the utilization of diffusion probability models in the realm of meta-atoms design. In subsequent research, we plan to harness the exceptional conditional accuracy inherent in diffusion probability models to develop and fabricate intricate freeform metasurfaces, such as those required for complex applications in real-world scenarios.

\medskip
\noindent \textbf{Supporting Information} \par 
\noindent Supporting Information is available from the author.

\medskip
\noindent \textbf{Acknowledgements}\par 
\noindent This research is supported by the National Key Research and Development Program (2020YFB1806400 and 2020YFB1806405).

\medskip
\noindent \textbf{Conflict of Interest}\par 
\noindent The authors declare no conflicts of interest.

\medskip
%
\bibliographystyle{MSP}
\bibliography{ref}


\end{document}